\documentclass[10pt,a4paper,twocolumn]{article}

\pdfoutput=1

\usepackage[T1]{fontenc}
\usepackage[utf8]{inputenc}

\usepackage{times}
\usepackage{multicol}
\usepackage{graphicx}
\usepackage{amsmath}
\usepackage{url}
\usepackage[hidelinks,
            pdfauthor={Lenz Furrer, Joseph Cornelius, Fabio Rinaldi},
            pdftitle={Parallel sequence tagging for concept recognition}]{hyperref}

\newcommand{\eg}{e.\,g.\ }
\newcommand{\ie}{i.\,e.\ }

\begin{document}

\title{\textbf{Parallel sequence tagging for concept recognition}}
\author{\textbf{Lenz Furrer\,$^\text{1,3}$, Joseph Cornelius\,$^\text{1}$ and Fabio Rinaldi\,$^\text{1,2,3}$}\\
$^\text{1}$Department of Computational Linguistics, University of Zurich, Switzerland\\
$^\text{2}$Dalle Molle Institute for Artificial Intelligence Research (IDSIA), Switzerland\\
$^\text{3}$Swiss Institute of Bioinformatics, Switzerland}
\date{}

\maketitle

\begin{abstract}
\textbf{Background:} Named Entity Recognition (NER) and Normalisation (NEN) are core components of any text-mining system for biomedical texts.
In a traditional concept-recognition pipeline, these tasks are combined in a serial way, which is inherently prone to error propagation from NER to NEN.
We propose a parallel architecture, where both NER and NEN are modeled as a sequence-labeling task, operating directly on the source text.
We examine different harmonisation strategies for merging the predictions of the two classifiers into a single output sequence.\\
\textbf{Results:} We test our approach on the recent Version~4 of the CRAFT corpus.
In all 20 annotation sets of the concept-annotation task, our system outperforms the pipeline system reported as a baseline in the CRAFT shared task 2019.\\
\textbf{Conclusions:} Our analysis shows that the strengths of the two classifiers can be combined in a fruitful way.
However, prediction harmonisation requires individual calibration on a development set for each annotation set.
This allows achieving a good trade-off between established knowledge (training set) and novel information (unseen concepts).\\
\textbf{Availability and Implementation:} Source code freely available for download at \url{https://github.com/OntoGene/craft-st}. Supplementary data are available at arXiv online.\\
\textbf{Contact:} \href{rinaldi@cl.uzh.ch}{rinaldi@cl.uzh.ch}
\end{abstract}

\section*{Background}

Concept recognition is a fundamental task in text mining for biomedical texts.
Biomedical text mining finds applications in literature analysis, literature-based discovery but also over other types of text, such as clinical records and social media.
For most applications, identifying occurrences of biomedical concepts is an essential first step.
The task is usually tackled in a two-stage approach:
First, named entity recognition (NER), or span detection, is concerned with identifying textual mentions of relevant entities, such as proteins, chemicals, or species.
Second, the identified mentions are assigned to a concept entry in a controlled vocabulary, which is referred to as named entity normalisation (NEN), linking, or grounding.
Typically, the two steps are performed in a sequential manner, using a sequence classifier for NER and a ranking- or rule-based module for NEN.
While this approach allows focusing on different methods for the individual steps, it suffers from error propagation, an inherent drawback of any pipeline architecture.
For example, a certain NEN system might have excellent accuracy when using ground-truth spans as input, but its performance will decrease when operating on the imperfect output of a span tagger.
In particular, a normaliser might be inclined or even forced to predict a concept ID for spurious spans, and it cannot recover from cases where a span is missing.

In this work, we investigate an alternative architecture for concept recognition, which alleviates the problem of error propagation: parallel sequence tagging for NER and NEN.
In this architecture, NEN is modeled as a sequence-classification problem (like NER) and applied to the input text independently of the span tagger.
The predictions of the two taggers are harmonised using different strategies, the choice of which is a hyperparameter of the complete system.
We test our approach with a manually annotated dataset for biomedical concepts, the CRAFT corpus, continuing the efforts from our participation in the CRAFT shared task 2019.

\subsection*{Related Work}

Concept recognition has often been approached as a pipeline of NER+NEN.
For NER, sequence labeling with conditional random fields (CRF) has dominated the field to present, be it pure CRF as in Gimli \cite{campos-et-al:2013:gimli} or DTMiner \cite{xu-et-al:2016}, on top of a recurrent neural network as in HUNER \cite{weber-et-al:2019}, Saber \cite{giorgi-bader:2019}, or DTranNER \cite{hong-lee:2020}, or even as the head of a BERT-based system as in SciBERT \cite{beltagy2019scibert}.
BERN \cite{kim2019neural} performs NER by fine-tuning BioBERT alone, even though \cite{yu2019bertNER} report improved results when stacking CRF atop BioBERT.
Different approaches have been taken to NEN, where extracted mentions are mapped to a vocabulary: exact match as in Neji \cite{campos-et-al:2013:neji}, expert-written rules \cite{dsouza-ng:2015:ACL-IJCNLP}, learning-to-rank as in DNorm \cite{leaman-et-al:2013}, or sequence-to-sequence prediction \cite{hailu-et-al:2019:bioRxiv}.

Knowledge-based concept-recognition systems like Jensen tagger \cite{pletscher-frankild-jensen:2019} or NOBLE coder \cite{tseytlin-et-al:2016} do not allow for a clear separation between NER and NEN, as span detection and linking happens at once, even if machine-learning components are added for improving accuracy, like for OGER++ \cite{furrer-et-al:2019} or RysannMD \cite{cuzzola-et-al:2017}.
Joint approaches like TaggerOne \cite{leaman-lu:2016}, JLink \cite{ter-horst-et-al:2017:LDK}, and others \cite{lou-et-al:2017,zhao-et-al:2019:AAAI} however, have separate modules for NER and NEN, which are trained simultaneously.
The multi-task sequence labeling architecture for NER and NEN in \cite{zhao-et-al:2019:AAAI} has been highly inspirational for the present work, although we were unable to reproduce their results, even using the code that the authors made publicly available.

\subsection*{CRAFT corpus and shared task}

The Colorado Richly Annotated Full-Text (CRAFT) corpus \cite{bada-et-al:2012,cohen-et-al:2017} is a collection of 97 scientific articles from the biomedical domain.
It is manually annotated for syntactic structure, coreferences, and bio-concepts (entities), the last of which are used in the present study.
In the latest release (Version~4), the concept annotations are divided into 10 sets of different entity types, which are provided in two versions each (proper and extended%
\footnote{The extended annotations are based on a modified version of the reference ontologies. Through these modifications, the corpus creators aimed at more accurately capturing language use in scientific literature.}%
), for a total of 20 separate annotation sets over the same text collection.
The concepts are linked to 8 different ontologies, as shown below (ontology in parentheses):
\begin{itemize}
  \setlength{\itemsep}{0mm}
  \item[] \textbf{CHEBI:} chemicals/small molecules (Chemical Entities of Biological Interest)
  \item[] \textbf{CL:} cell types (Cell Ontology)
  \item[] \textbf{GO\_CC:} cellular and extracellular components and regions (Gene Ontology)
  \item[] \textbf{GO\_BP:} biological processes (Gene Ontology)
  \item[] \textbf{GO\_MF:} molecular functionalities possessed by genes (Gene Ontology)
  \item[] \textbf{MOP:} chemical reactions and other molecular processes (Molecular Process Ontology)
  \item[] \textbf{NCBITaxon:} biological taxa and organisms (NCBI Taxonomy)
  \item[] \textbf{PR:} proteins, genes, and transcripts (Protein Ontology)
  \item[] \textbf{SO:} biomacromolecular entities, sequence features (Sequence Ontology)
  \item[] \textbf{UBERON:} anatomical entities (UBERON)
\end{itemize}
The extended annotations are referred to by appending EXT to the abbreviations for the proper annotations (CHEBI\_EXT, CL\_EXT etc.).

The CRAFT corpus has been used in a range of studies.
Through repeated improvements and extensions over time, the corpus has become a high-quality resource with rich annotations, but it also led to the situation that most experiments are not directly comparable to each other, as their setup differs in many ways.
In the first release of the CRAFT corpus, only 67 articles were available.
The remaining 30 documents were not released until the evaluation period of the CRAFT shared task \cite{baumgartner-et-al:2019:BioNLP-OST}, where they served as a test set, and the subsequent release of Version~4.
Most prior work was carried out with an older version of CRAFT,
\ie using a different test set, possibly an earlier stage of annotations and a different evaluation method, which means that results are not directly comparable.

While the majority of studies is concerned with concept recognition (\ie systems that predict IDs), some are restricted to NER, \eg \cite{verspoor-et-al:2012,crichton-et-al:2017,giorgi-bader:2019}.
Methodologically, the approaches range from pure dictionary-based \cite{groza-verspoor:2015,tseytlin-et-al:2016} to entirely example-based systems \cite{hailu:2019}, even though the NEN step almost always includes dictionary lookup.
Since no official test set was available prior to Version~4, many experiments use an arbitrary train/test split \cite{basaldella-et-al:2017} or apply evaluation to the entire corpus \cite{campos-et-al:2013:neji}.
The metrics used are consistently precision, recall and F-score, but differences exist with respect to considering partial matches.
Also, many studies do not cover the full set of annotations, but rather focus on a small selection of entity types, such as Gene Ontology \cite{yang-chiang:2018} or gene mentions \cite{verspoor-et-al:2012}.

\section*{Methods}

We propose a paradigm for biomedical concept recognition where named entity recognition (NER) and normalisation (NEN) are tackled in parallel.
In a traditional NER+NEN pipeline, the NEN module is restricted to predict concept labels (IDs) for the spans identified by the NER tagger.
In order to avoid the error propagation inherent to this serial approach, we drop this restriction and provide the full input sequence to the normaliser.
As such, we cast the normalisation task as a sequence-tagging problem -- very much like an NER tagger, but with a considerably larger tag set, consisting of all concept IDs of the training data.

\subsection*{Design implications}

\begin{figure}[tb]
\centerline{\includegraphics[width=\columnwidth,trim=30 20 20 40,clip]{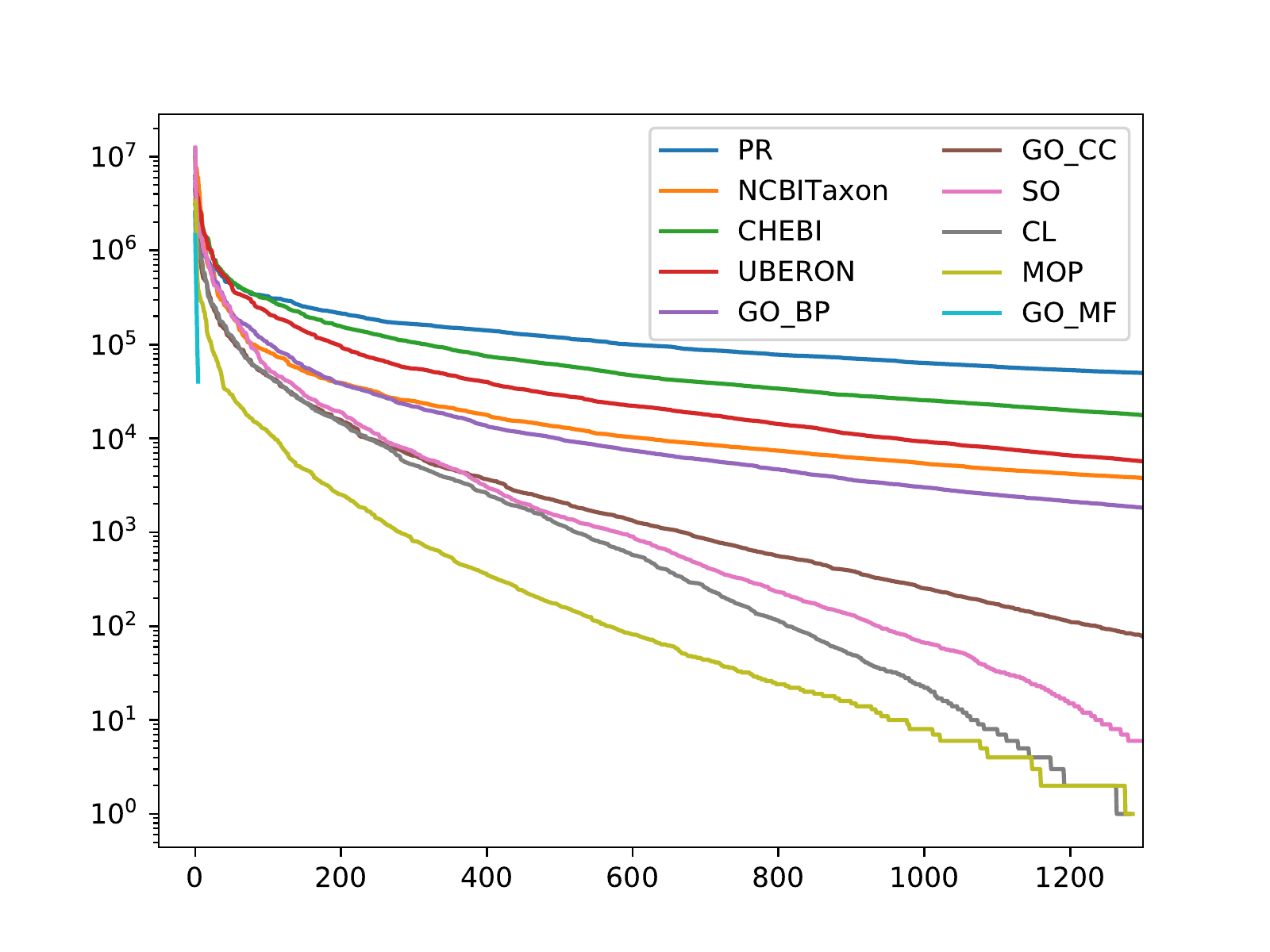}}
\caption{%
  Occurrence counts (y axis, log scale) of the most frequent bio entities in a large subset of PubMed, ordered by their rank (x axis).
  The documents were automatically annotated by a dictionary-based tagger (OGER).
  High-frequency false-positives were manually removed.
  The plot shows that a small number of frequent entities accounts for a majority of the occurring mentions, resembling a Zipfian distribution (see also \cite[p.~569]{boguslav-et-al:2018:PSB}).}
\label{fig:zipf}
\end{figure}

Modeling concept normalisation as sequence tagging has a number of drawbacks.
As discussed in the next section, the CoNLL representation of the data enforces exactly one label for each token, which disallows learning and predicting annotations with overlapping and discontinuous spans.
This representation also entails that the model has to produce a consistent series of individual predictions in order to correctly label a multi-word expression.
This often means that highly ambiguous tokens like prepositions, numbers, or single letters must be interpreted correctly in context (\eg ``of'' in ``inhibitor of calpain'', ``I'' in ``hexokinase I'').
As the most serious limitation, a sequence tagger can only ever predict labels it has seen during training, which restricts the label set of the trained system to a fraction of the target label set (the ontology) in many cases.
Since many concepts occur extremely rarely in the biomedical literature (cf.\ Figure~\ref{fig:zipf}), this limitation might not critically reduce performance measured on a typical evaluation data set.
However, it is highly undesirable to have a tagger that is completely incapable of predicting labels beyond the training set.

On the other hand, the ID-tagging architecture is technically an end-to-end concept-recognition system, \ie it does not depend on any span predictions, which means that the NER step could potentially be skipped entirely.
However, due to the small number of tags, span tagging is far more robust with respect to ambiguous tokens and unseen concepts.
By adding span predictions, we might thus be able to overcome the limitations of direct ID tagging.
Therefore, we chose to combine the strengths of span and ID tagging by applying both in parallel and merging the results in postprocessing.

\subsection*{Data preparation}

\begin{figure}[tb]
\centerline{\includegraphics[width=\columnwidth]{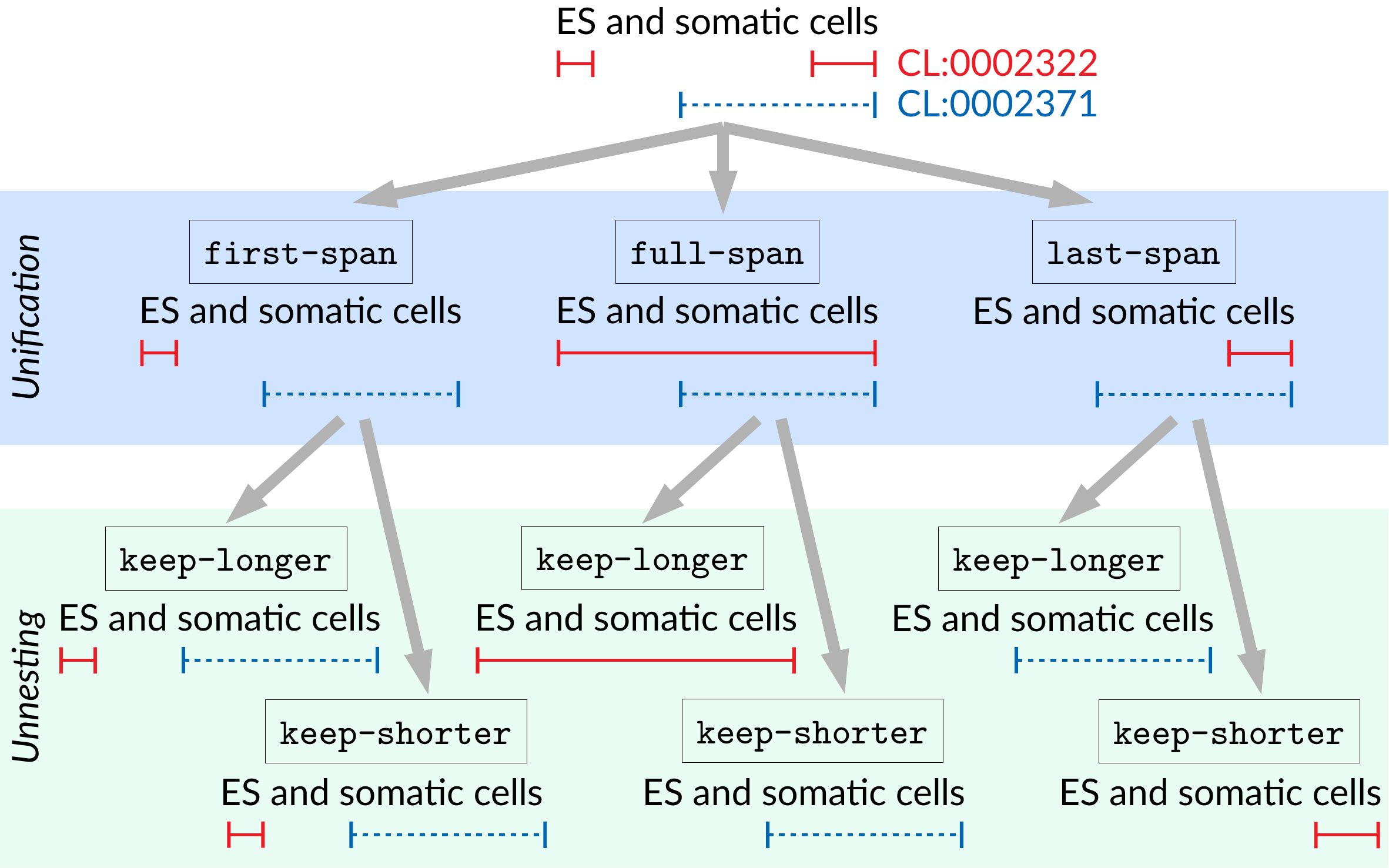}}
\caption{%
  Example phrase with a discontinuous annotation (``ES \dots\ cells'', solid red spans) that partially overlaps with a contiguous annotation (``somatic cells'', dashed blue spans).
  The annotations are simplified in two steps (unification and unnesting), for which different strategies are compared.
  In this example, the six possible combinations produce four different outcomes, of which three have lost one annotation entirely.}
\label{fig:interlaced}
\end{figure}

Our system processes documents in a variant of the CoNLL format, \ie a verticalised format where each text token is assigned exactly one label.
Based on our architecture with two sequence classifiers, we employed two different label sets.
For the span tagger, the text is tagged with IOBES labels, \ie each token is assigned one of the five labels I, O, B, E, or S.
Entities spanning only a single token are annotated with S.
For multi-word entities, the first and last token are tagged with B and E, respectively, and any intervening tokens with I.
The rest of the text (\ie all tokens outside of an entity) are annotated with O.
For the ID tagger, all tokens of an entity are tagged with the respective concept ID.
We added a NIL label to mark non-entity tokens, analogously to the O tag of the span tagger.

This representation does not have the same expressiveness as the stand-off format used in CRAFT, which offers great flexibility for anchoring annotations in the text.
In particular, the CRAFT corpus contains discontinuous annotations (multiple non-adjacent text spans for the same annotation), overlapping annotations (words shared by multiple annotations) and sub-word spans (annotation refers to part of a word).
Since these complex annotations cannot be represented with token-level labels, their structure needs to be simplified.

In order to measure the performance impact of this simplification, we converted the reference annotations of the training set to CoNLL format and back to stand-off using the \emph{standoff2conll} suite \cite{url:standoff2conll}.
This utility offers two strategies for unifying discontinuous annotations (\emph{full-span} and \emph{last-span}), to which we added a third option (\emph{first-span}) \cite{url:standoff2conll-fork}.
For unnesting overlapping annotations, two strategies are available as well (\emph{keep-longer} and \emph{keep-shorter}).
The effect of unifying and unnesting annotations is illustrated in Figure~\ref{fig:interlaced}.
Sub-word annotations are extended to span entire tokens.

After this round-trip conversion, the annotations are run through the official evaluation suite provided by CRAFT \cite{url:craft-eval-suite}.
Table~\ref{tab:roundtripconv} shows the results for different combinations of unification and unnesting strategies on the non-extended annotation sets.
These numbers mark the upper limit for a system trained on input data in CoNLL format.
For all annotation sets, using the \emph{first-span} and \emph{keep-longer} strategies achieved the highest F-score.

\begin{table}[tb]
\caption{Upper bound of annotation performance (F-score) when using the CoNLL format, comparing different simplification strategies.}
\label{tab:roundtripconv}
\small\setlength{\tabcolsep}{2pt}
\begin{center}
\begin{tabular}{lcccc}
\hline
Unification & \multicolumn{2}{c}{first-span}  & full-span       & last-span \\
Unnesting & keep-shorter    & keep-longer & \multicolumn{2}{c}{keep-longer} \\
\hline
CHEBI     &         0.9979  & \textbf{0.9980} &         0.9972  &         0.9974 \\
CL        &         0.9706  & \textbf{0.9720} &         0.9574  &         0.9692 \\
GO\_BP    &         0.9607  & \textbf{0.9626} &         0.9587  &         0.9570 \\
GO\_CC    &         0.9811  & \textbf{0.9813} &         0.9801  &         0.9785 \\
GO\_MF    & \textbf{0.9974} & \textbf{0.9974} & \textbf{0.9974} & \textbf{0.9974} \\
MOP       & \textbf{0.9967} & \textbf{0.9967} & \textbf{0.9967} & \textbf{0.9967} \\
NCBITaxon & \textbf{0.9996} & \textbf{0.9996} &         0.9995  & \textbf{0.9996} \\
PR        &         0.9624  & \textbf{0.9627} &         0.9619  &         0.9618 \\
SO        &         0.9829  & \textbf{0.9831} &         0.9816  &         0.9825 \\
UBERON    &         0.9792  & \textbf{0.9798} &         0.9776  &         0.9780 \\
\hline
\end{tabular}
\end{center}
\small
Note: for reasons of clarity, the combinations \emph{full-span/keep-\hspace{0cm}shorter} and \emph{last-span/keep-shorter} are omitted; their results are in most cases inferior to those presented in this table.
\end{table}

\subsection*{Architecture}

The sequence taggers used in our experiments are built atop a pretrained language-representation model, BioBERT \cite{lee2019biobert}, which in turn extends BERT \cite{devlin2019bert}.
BERT is an attention-based multi-layer neural network which learns context-dependent word-vector representations.
It creates bidirectional contextual representations of a token from unlabeled text conditioned on the left and the right context.
BERT is trained to solve two tasks: first, to predict whether two sentences follow each other, and second, to predict a randomly masked token from its context.
After a slight modification to its architecture, training of BERT can be continued on a different task like NER; this process is referred to as fine-tuning with a task-specific head.

For our experiments, we downloaded BioBERT v1.1, which includes code, configuration and pretrained parameters.
BioBERT is based on BERT$_\text{BASE}$, which was pretrained for 1M steps by Devlin et al.\ \cite{devlin2019bert} on a 3.3B-word corpus from the general domain (English Wikipedia, BooksCorpus).
Lee et al.\ \cite{lee2019biobert} continued training for another 1M steps on a 4.5B-word biomedical corpus (PubMed abstracts).
Finally, we fine-tuned BioBERT for sequence-tagging on the CRAFT corpus for 55~epochs (approximately 53k steps).

To perform NER and NEN in parallel, we used two different tag sets for fine-tuning, as described in the previous section:
IOBES labels for the span tagger and the set of all concept IDs for the ID tagger.
In addition to that, both taggers used a small set of tags inherited from the original BERT implementation, which flag tokens with a special function, such as padding, sub-word unit and sentence boundary.
We trained a pair of span and ID tagger for each annotation set, which resulted in a total of 40 individual models.

The predictions of the span tagger are always aligned with the IDs produced by a dictionary-based concept-recognition system, OGER \cite{furrer-rinaldi:2017:BC5.5,furrer-et-al:2019}.
OGER detects mentions of ontology terms in running text through efficient fuzzy-matching.
We manually optimised OGER's configuration on the CRAFT training set.
We used no additional terminology resources besides the ontologies provided with the corpus.
However, we manually added a handful synonyms for GO\_MF.
This combined system resembles a classical NER+NEN pipeline, where the high-recall output of the dictionary-based system is combined with the context-aware span detection using an example based classification model.

\subsection*{Hyperparameter tuning}

In order to determine the best hyperparameters for each annotation set, we performed extensive grid search in cross-validation over the training set.
In particular, we investigated the following configurations:
\begin{itemize}
  \item[] \textbf{ontology pretraining:} enable/disable
  \item[] \textbf{abbreviation expansion:} enable/disable
  \item[] \textbf{prediction harmonisation:} 6 strategies
\end{itemize}
If ontology pretraining is enabled, the ID classifier is trained on synonym--ID pairs from the terminology for 20 epochs before switching to the actual training corpus.
For abbreviation expansion, we first used Ab3P \cite{sohn-et-al:2008} to detect abbreviation definitions, then replaced occurrences of short forms with the corresponding long form.
For harmonising the predictions of the two classifiers, we compared six different strategies; these are described in the next section.

From previous experiments \cite{furrer-et-al:2019:BioNLP-OST}, we knew that ontology pretraining has a positive effect for some, but a negative effect for other annotation sets.
We therefore concluded that hyperparameters had to be tuned individually for each of the 20 annotation sets.
In order to obtain reliable figures, we performed 6-fold cross-validation with up to 3 runs for each combination.

As we expected, ontology pretraining yielded a mixed picture.
In many cases, a clear decision was not possible, as repeated runs gave contradictory results.
Unexpectedly, abbreviation expansion showed a clear improvement only for CL and a slight improvement for GO\_MF; in all other cases (including CL\_EXT and GO\_MF\_EXT) the results decreased.
We decided to disable both ontology pretraining and abbreviation expansion, as the isolated merits do not justify the added complexity.

For prediction harmonisation, the best strategy for each annotation set is given in Table~\ref{tab:best-harmonisation} and discussed in the following section.
The full results for the whole tuning phase are included in the ancillary files.

\subsection*{Harmonising predictions}

The predictions of the span and ID classifier are not guaranteed to agree, even if trained jointly.
Disagreement occurs if the span classifier predicts a relevant tag (B, I, E, S) for a particular token while the ID classifier predicts NIL, or, conversely, if the ID classifier predicts a specific concept for a token tagged as irrelevant (O) by the span classifier.
In addition, the dictionary feature of the knowledge-based entity recogniser might or might not agree with the neural predictions.
This results in $2\times2\times2=8$ prediction patterns concerning the relevance of a given token.

\begin{figure}[tb]
\centerline{\includegraphics[width=\columnwidth]{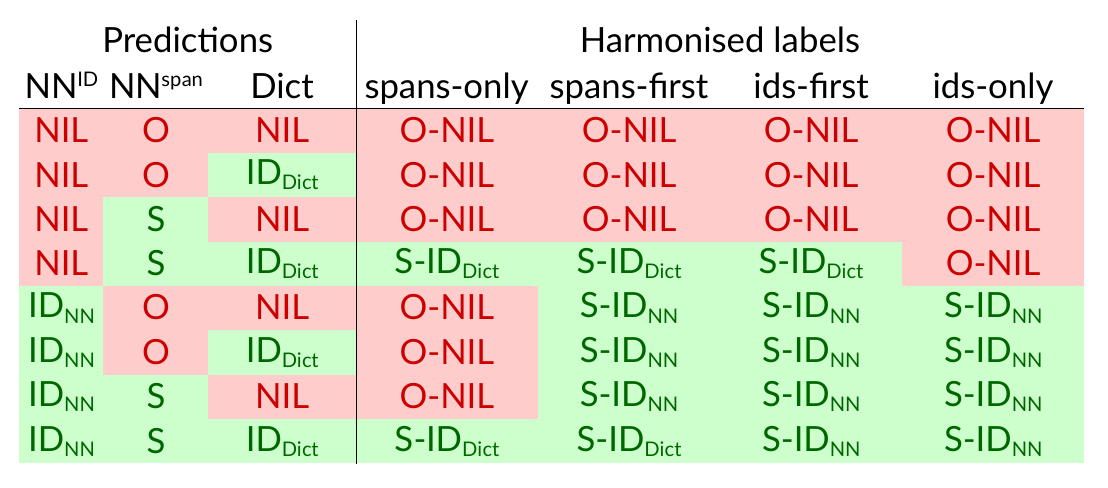}}
\caption{%
  Strategies for harmonising predictions of the ID classifier (NN$^\text{ID}$), span tagger (NN$^\text{span}$), and dictionary-based entity recogniser (Dict).
  ``S'' is a cover symbol representing any relevant tag (B, I, E, S); ``ID$_\text{NN}$'' and ``ID$_\text{Dict}$'' refer to any non-NIL prediction of the ID classifier and entity recogniser, respectively.}
\label{fig:harmonisation}
\end{figure}

\begin{figure}[tb]
\centerline{\includegraphics[width=\columnwidth]{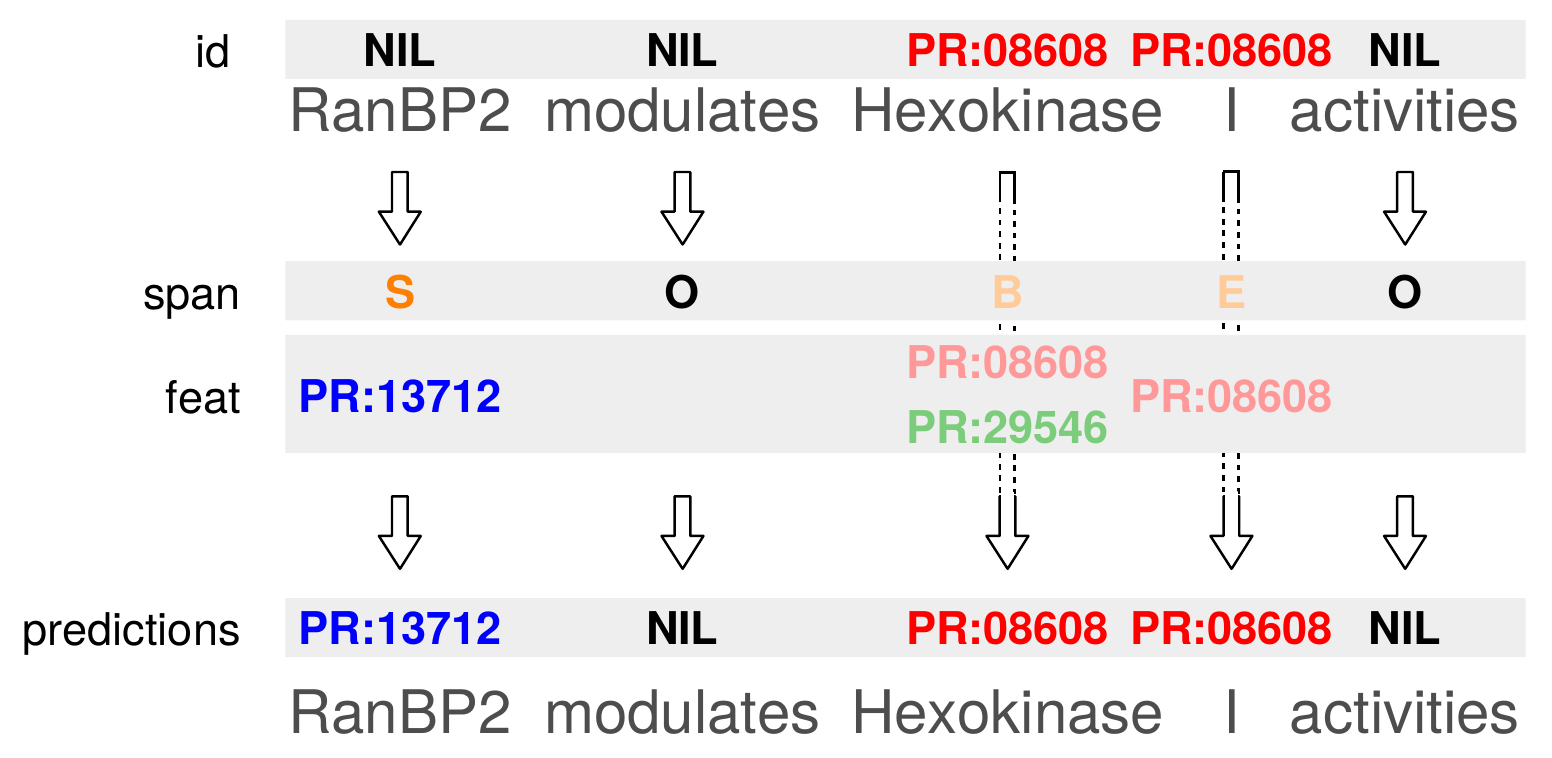}}
\caption{%
  Predictions for PR on a short phrase, harmonised with the \emph{ids-first} strategy.
  Using the \emph{spans-only} or \emph{spans-first} strategy would yield the same result in this example, since the ID and span predictions are identical for ``Hexokinase I''.}
\label{fig:backoff}
\end{figure}

We considered four different strategies for harmonising conflicting predictions: \emph{spans-only}, \emph{ids-only}, \emph{spans-first}, and \emph{ids-first} (cf.\ Figure~\ref{fig:harmonisation}).
These strategies are heuristics with a predetermined bias towards one of the two classifiers.
Two additional strategies (\emph{mutual} and \emph{override}), which use the confidence scores for balancing the classifiers, consistently produced worse results compared to the simpler bias strategies.
The score-based strategies are thus not discussed here; however, we used and described the \emph{mutual} strategy when participating in the CRAFT shared task \cite[p.~188]{furrer-et-al:2019:BioNLP-OST}.
The systematic application of different harmonisation strategies is one of the major differences of this work compared to the work presented at the shared-task workshop.

With the \emph{spans-only} strategy, the ID predictions are completely ignored.
In order to provide a concept label, the span predictions are combined with the dictionary feature provided by OGER; in case of multiple features, an arbitrary decision is taken (lexically lowest ID).
Since a concept label is always required, span predictions without a supporting feature have to be dropped.

With the \emph{ids-only} strategy, the predictions are based primarily on the ID predictions, whereas the span predictions are overridden (\eg the span tag cannot be O when the ID classifier predicts a non-NIL concept).
The dictionary feature is ignored in the decision.

The \emph{spans-first} and \emph{ids-first} strategies are combinations of the previous two.
With the former, the \emph{spans-only} strategy is applied first, backing off to the \emph{ids-only} strategy if the outcome is O-NIL.
Analogously, the \emph{ids-first} strategy gives preference to \emph{ids-only}.
An example with partially disagreeing predictions is given in Figure~\ref{fig:backoff}.

We compared the effect of the different strategies in a 6-fold cross-validation over the training set.
For each annotation set, we determined the best harmonisation strategy based on F-score according to the official evaluation suite.
As shown in Table~\ref{tab:best-harmonisation}, using both span and ID predictions was beneficiary most of the time.
In many cases, the same strategy worked best for the proper and extended classes.
Intuitively, the choice of \emph{spans-only} for proteins makes sense, as PR[\_EXT] shows an exceedingly high number of different concepts with a small overlap between training and test data, which is a tough scenario for the ID tagger.
Conversely, entity types with a limited number of distinct concepts in the corpus like sequences and organisms rely more heavily on the ID tagger.
The choice of harmonisation strategy was fixed as a hyperparameter for the test-set predictions.

\begin{table}[tb]
\caption{Best-performing harmonisation strategy by annotation set, based on 6-fold cross-validation over the training set.}
\label{tab:best-harmonisation}
\begin{center}
\begin{tabular}{lcc}
\hline
Annotation set & proper      & extended \\
\hline
CHEBI     & spans-first      & spans-first \\
CL        & spans-first      & ids-only \\
GO\_BP    & spans-first      & ids-only \\
GO\_CC    & spans-first      & spans-first \\
GO\_MF    & spans-/ids-first & spans-first \\
MOP       & spans-/ids-first & ids-first \\
NCBITaxon & ids-first        & ids-first \\
PR        & spans-only       & spans-only \\
SO        & ids-only         & ids-only \\
UBERON    & spans-first      & spans-first \\
\hline
\end{tabular}
\end{center}
\small
% Ranking by slot error rate (SER) and F-score yielded the same winner in all cases.
For GO\_MF and MOP, results for \emph{spans-first} and \emph{ids-first} are identical.
\end{table}

\section*{Results and discussion}

\begin{table*}[tbp]
\caption{Results for our current BioBERT system, best system reported in the shared-task paper \protect{\cite{furrer-et-al:2019:BioNLP-OST}}, and the official baseline.}
\label{tab:results}
\small
\begin{center}
\begin{tabular}{llllll}
\hline
Annotation set & System & \multicolumn{2}{c}{proper} & \multicolumn{2}{c}{extended} \\
          &             &  SER  & F-score &  SER   & F-score \\
\hline
          & baseline    & 0.44   & 0.72   & 0.29   & 0.80   \\
CHEBI     & shared-task & 0.3388 & 0.7700 & 0.2571 & 0.8209 \\
          & current     & 0.2492 & 0.8528 & 0.2289 & 0.8459 \\
\hline
          & baseline    & 0.53   & 0.61   & 0.33   & 0.73   \\
CL        & shared-task & 0.4862 & 0.6657 & 0.3361 & 0.7484 \\
          & current     & 0.4013 & 0.7526 & 0.2777 & 0.7926 \\
\hline
          & baseline    & 0.39   & 0.72   & 0.29   & 0.79   \\
GO\_BP    & shared-task & 0.3047 & 0.8037 & 0.2786 & 0.8138 \\
          & current     & 0.2587 & 0.8297 & 0.2015 & 0.8506 \\
\hline
          & baseline    & 0.44   & 0.71   & 0.20   & 0.88   \\
GO\_CC    & shared-task & 0.3788 & 0.7645 & 0.1678 & 0.8936 \\
          & current     & 0.2817 & 0.8219 & 0.1486 & 0.9073 \\
\hline
          & baseline    & 0.07   & 0.95   & 0.45   & 0.66   \\
GO\_MF    & shared-task & 0.0319 & 0.9838 & 0.3881 & 0.7438 \\
          & current     & 0.0149 & 0.9904 & 0.4135 & 0.7139 \\
\hline
          & baseline    & 0.43   & 0.75   & 0.36   & 0.79   \\
MOP       & shared-task & 0.2684 & 0.8705 & 0.3080 & 0.8437 \\
          & current     & 0.1567 & 0.9188 & 0.1713 & 0.9082 \\
\hline
          & baseline    & 0.07   & 0.96   & 0.07   & 0.96   \\
NCBITaxon & shared-task & 0.0537 & 0.9694 & 0.0466 & 0.9722 \\
          & current     & 0.0436 & 0.9744 & 0.0460 & 0.9704 \\
\hline
          & baseline    & 0.69   & 0.48   & 0.62   & 0.52   \\
PR        & shared-task & 0.3052 & 0.8026 & 0.3030 & 0.8011 \\
          & current     & 0.3068 & 0.8041 & 0.3130 & 0.7951 \\
\hline
          & baseline    & 0.21   & 0.86   & 0.18   & 0.89   \\
SO        & shared-task & 0.1593 & 0.9027 & 0.1230 & 0.9187 \\
          & current     & 0.1206 & 0.9223 & 0.0899 & 0.9419 \\
\hline
          & baseline    & 0.41   & 0.70   & 0.36   & 0.75   \\
UBERON    & shared-task & 0.3752 & 0.7488 & 0.3371 & 0.7714 \\
          & current     & 0.2790 & 0.8177 & 0.2537 & 0.8315 \\
\hline
\end{tabular}
\end{center}
\small
In case of the shared-task systems, the results were selected independently for SER and F-score, \ie the two scores for a given annotation set do not necessarily come from the same system.
For the baseline and the current BioBERT system, however, only one system was evaluated for each annotation set.
\end{table*}

We evaluated our concept-recognition system using the official evaluation suite \cite{url:craft-eval-suite}.
Performance is measured in terms of F-score, \ie the harmonic mean of precision and recall, and slot error rate (SER) \cite{makhoul-et-al:1999:DARPA}.
Both metrics are based on the counts of matches (true positives), substitutions (partial errors), insertions (false positives), and deletions (false negatives).
Partially correct predictions are assigned a similarity score $m$ in the range $[0, 1]$, which measures the accurateness of the predicted spans and concept labels \cite{bossy-et-al:2013:BioNLP-ST}.
The similarity score incorporates a notion of textual overlap (Jaccard index at the character level) and a weighted measure of shared ancestors in the ontology hierarchy, as introduced in \cite{wang-et-al:2007}.
The fractional value $m$ is added to the match count, whereas the remainder $1-m$ is counted as a substitution.
While precision, recall, and F-score are figures of merit ranging from 0 (worst) to 1 (best), SER is a measure of error that assigns 0 to a perfect system and higher values to lower performance.
Even though the values for SER and F-score often correlate, they are not guaranteed to produce identical rankings.
In particular, SER is more sensitive to false-positive errors than F-score, and low precision has a stronger impact on SER than low recall.

The results for our parallel NER+NEN system are given in Table~\ref{tab:results}.
The scores are compared to our systems developed for the shared task \cite{furrer-et-al:2019:BioNLP-OST} and to the official baseline published in the workshop overview \cite{baumgartner-et-al:2019:BioNLP-OST}.
Our system consistently achieves better scores than the baseline, which is a pipeline with a CRF-based span tagger and a BiLSTM-based concept classifier that were also trained on the CRAFT corpus alone.
For most annotation sets, our current system performed better than the best system presented in the shared-task paper, with the exception of GO\_MF\_EXT and PR\_EXT.
For NCBITaxon\_EXT and PR, the comparison is inconclusive, as SER and F-score give contradictory rankings.

\subsection*{Effect of harmonisation}

\begin{figure*}[tb]
\centerline{\includegraphics[width=\textwidth]{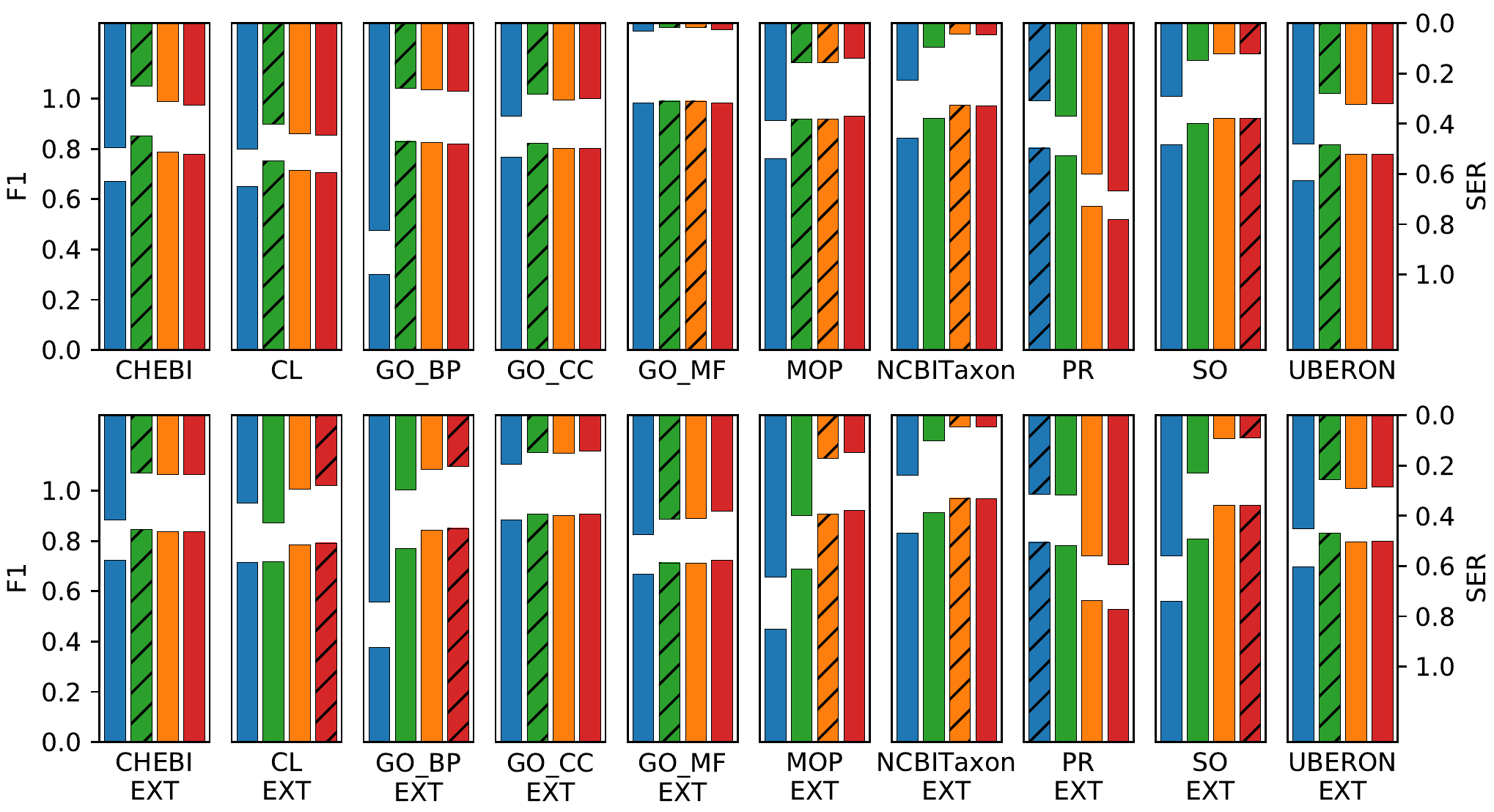}}
\centerline{\includegraphics[width=\textwidth]{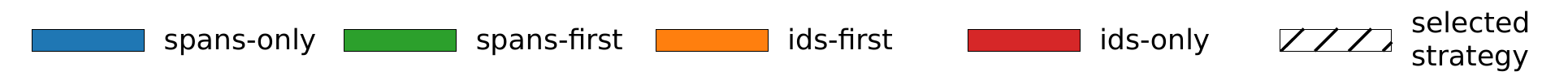}}
\caption{
  F1 and SER scores for different harmonisation strategies.
  Top-down bars represent SER (right scale), bottom-up bars represent F-score (left scale).
  Hatched bars denote the strategies used in the final results, as determined through hyperparameter tuning.
  For GO\_MF and MOP, \emph{spans-first} and \emph{ids-first} yielded identical results in the training set, which was repeated in the test set experiments.
  The exact figures are available as a table in the ancillary files.}
\label{fig:ablation}
\end{figure*}

In order to measure the effect of the different harmonisation strategies, we evaluated all four strategies on the test set, as shown in Figure~\ref{fig:ablation}.
This study also serves as a validation for our hyperparameter-tuning approach, \ie whether cross-validation on the training set can be used for reliably picking the best-suited harmonisation strategy.
For the majority of the annotation sets, the picked strategy also worked best for the test set.
Where the picked strategy was not the best (GO\_MF\_EXT, MOP[\_EXT]), the difference to the top-performing strategy was comparatively small.

\subsection*{Unseen concepts}

As stated above, a major limitation of trained sequence labeling for IDs is the inability to predict concepts not seen among the training examples.
An important goal of combining the ID tagger with a span tagger and dictionary-based predictions is to overcome this limitation.
To study the effect of the different harmonisation strategies on unseen concepts, we performed another evaluation on a subset of the annotations.
To this end, we filtered both ground truth and predictions of the test set to contain only annotations with concept labels that are not used in the training set.

Table~\ref{tab:unseen} shows precision and recall scores as well as annotation counts for the subset of unseen concepts.
The \emph{ids-only} strategy is omitted in the table, as this configuration can never predict unseen concepts.
The \emph{spans-only} and \emph{spans-first} strategies systematically yield identical results, as they only differ in cases where the latter backs off to ID predictions, which have been filtered out in this evaluation.
With the \emph{ids-first} strategy, many span predictions for unseen concepts are shadowed by an ID prediction for a concept known from the training set (which is then ignored in this specific evaluation).
For some annotation types (\eg CHEBI[\_EXT], GO\_BP[\_EXT], SO[\_EXT]), the removal of known concepts improves precision, \ie more false positives than true positives were removed.
In other cases, precision suffers from the removal.
Recall decreases in all cases, as is to be expected for an evaluation that focuses on more difficult examples.

\begin{table*}[tbp]
\caption{Precision and recall for unseen concepts in the test set.}
\label{tab:unseen}
\small
\begin{center}
\begin{tabular}{llrrccrrcc}
\hline
Annotation set & Harmonisation & \multicolumn{4}{c}{proper} & \multicolumn{4}{c}{extended} \\
& & \multicolumn{2}{c}{ref.\ count} & prec. & recall & \multicolumn{2}{c}{ref.\ count} & prec. & recall \\
& & unique & occ. & & & unique & occ. & & \\
\hline
CHEBI     & spans-only/-first & 110 &  447 & 0.7747 & 0.5199 & 134 &  538 & 0.6265 & 0.5462 \\
          & ids-first         &     &      & 0.8805 & 0.0867 &     &      & 0.7131 & 0.0530 \\
\hline
CL        & spans-only/-first &  52 &  484 & 0.8817 & 0.2222 &  52 &  484 & 0.6900 & 0.2338 \\
          & ids-first         &     &      & 0.8750 & 0.0723 &     &      & 0.4326 & 0.0152 \\
\hline
GO\_BP    & spans-only/-first & 120 &  484 & 0.6170 & 0.1402 & 126 &  508 & 0.2699 & 0.1621 \\
          & ids-first         &     &      & 0.7466 & 0.0524 &     &      & 0.4243 & 0.0175 \\
\hline
GO\_CC    & spans-only/-first &  32 &  184 & 0.6343 & 0.1965 &  36 &  231 & 0.5158 & 0.1853 \\
          & ids-first         &     &      & 0.4956 & 0.0458 &     &      & 0.3375 & 0.0058 \\
\hline
GO\_MF    & spans-only/-first &   1 &    1 &   --   &   --   &  73 &  416 & 0.5366 & 0.1393 \\
          & ids-first         &     &      &   --   &   --   &     &      & 0.5329 & 0.0090 \\
\hline
MOP       & spans-only/-first &   2 &    2 &   --   &   --   &   2 &    2 &   --   &   --   \\
          & ids-first         &     &      &   --   &   --   &     &      &   --   &   --   \\
\hline
NCBITaxon & spans-only/-first &  40 &   87 & 0.3805 & 0.2974 &  44 &   95 & 0.4070 & 0.3342 \\
          & ids-first         &     &      & 0.7346 & 0.1689 &     &      & 0.7363 & 0.1163 \\
\hline
PR        & spans-only/-first & 278 & 4782 & 0.8170 & 0.7350 & 309 & 5156 & 0.8286 & 0.7230 \\
          & ids-first         &     &      & 0.7402 & 0.1200 &     &      & 0.6852 & 0.0909 \\
\hline
SO        & spans-only/-first &  16 &  101 & 0.0962 & 0.0571 &  25 &  123 & 0.1586 & 0.2579 \\
          & ids-first         &     &      & 1.0000 & 0.0198 &     &      & 0.9345 & 0.1823 \\
\hline
UBERON    & spans-only/-first & 203 & 1297 & 0.7246 & 0.2447 & 207 & 1308 & 0.7142 & 0.2555 \\
          & ids-first         &     &      & 0.5913 & 0.0342 &     &      & 0.5342 & 0.0225 \\
\hline
\end{tabular}
\end{center}
\small
For each annotation set, the number of annotations (ref. count) in the test set is given, counting both occurrences (occ.) and unique labels (unique).
A dash for precision and recall means that the corresponding system did not predict any unseen concept at all (neither true nor false positive).
\end{table*}

\subsection*{Interpretation}

Tackling concept recognition for multiple entity types with a single architecture is very challenging, even if a separate model is trained for every annotation set.
The comparative results for the different harmonisation strategies (Figure~\ref{fig:ablation}) illustrate well how some annotation sets profit more from the span tagger (blue, left-most bars), others more from the ID tagger (red, right-most bars).
In many cases, merging predictions from the two taggers (middle bars) yields better results than relying on a single tagger (outer bars).
This preference does not directly correlate with ontology size: the two annotation sets with the largest ontologies (NCBITaxon and PR) show quite distinct result patterns.
However, it is possible to empirically determine how well each harmonisation strategy suits the characteristics of a given annotation set.
Using cross-validation over the training set resulted in robust estimations for ranking the harmonisation strategies.

The diversity of the individual annotation sets shows even more clearly when it comes to predicting unseen concepts.
In general, the level of precision and recall for unseen concepts varies greatly across annotation sets, as does the number of unseen concepts in the reference (cf.\ Table~\ref{tab:unseen}).
There is a loose negative correlation to the performance on the entire test set:
annotation sets like NCBITaxon[\_EXT] and SO[\_EXT] show high overall scores and low scores for unseen concepts, whereas more difficult sets like PR[\_EXT] have comparatively high precision and recall for unseen concepts.
A possible explanation is that the former annotation sets have little variability and a high overlap between training and test set, leading to a strong bias for known concepts (overfitting tendency), which is beneficiary for the test set as a whole, but not for the subset of unseen concepts.
The latter annotation sets show great variability of concept labels and surface names in the training data, which makes the task harder but also leads to better generalisation, as the classifier cannot achieve good performance by only learning a few frequent concepts.

\subsection*{Error analysis}

We performed an analysis of prediction errors in order to find potential weaknesses or systematic mistakes.
As expected, many errors are false negatives due to missing training examples.
There are several cases where spelled-out mentions are matched, whereas their abbreviated versions are missed.
For example, ``olfactory tubercle'' is correctly linked thanks to the dictionary-based predictions, while the ad-hoc acronym ``OT'' is missed.
False positive predictions are also frequently seen among abbreviations, which have an increased likelihood of being ambiguous.
For example, the short-hand ``NF'' denotes either ``neurofilament'' or ``nuclear factor'' in the training set, which cannot always be correctly distinguished by the classifier.

At first sight, it seems like abbreviation expansion should be able to alleviate errors like these.
Replacing short forms with their corresponding long forms increases chances for a dictionary match and, since it is performed within document scope, potentially reduces ambiguity.
However, abbreviation expansion is not guaranteed to work perfectly and can be a source of confusion even if it does.
For example, ``OT'' was correctly expanded to ``olfactory tubercle''.
Unfortunately, this misguided the classifier to label the term as \emph{olfactory bulb}, as the first token was only used for this concept in the training data.
In our experiments, the net effect of abbreviation expansion was negative, as stated above in the \emph{Hyperparameter tuning} section.

Sometimes, spurious predictions are caused by a substring shared with a training example.
Since the WordPiece tokeniser used in (Bio)BERT cuts unknown words into sub-word segments, the classifier sometimes associates a concept label with the fraction of a word, which might trigger false positives in unexpected contexts.
As an extreme example, mentions of ``PDGFR'', ``PFK'', ``PKD'', ``PI3K'', and ``PFKD'' are erroneously linked to \emph{phosphoglycerate kinase} (abbreviated ``PGK'').
This is most likely due to the shared initial letter, as the terms do not refer to semantically similar concepts (even though PFK and PI3K are also kinases).
Similarly, ``forkhead'' is linked to \emph{fork}, ``polymorphonuclear'' is linked to \emph{nucleus} and ``prosensory'' is linked to \emph{forebrain} (after the synonym ``prosencephalon'' seen in training data).

In some cases, the chosen harmonisation strategy prefers an erroneous label over a correct one.
For example, the term ``monkey'' is linked to \emph{mouse} by the ID tagger due to context (training: ``mouse kidney'', test: ``monkey kidney'').
Since the NCBITaxon systems are harmonised with the \emph{ids-first} strategy, this erroneous prediction overrides the correct annotation from the dictionary-based tagger.
Conversely, the dictionary predictions for ``insulin'' always link to PR:000009054, a specific protein.
In the ground truth, however, the more general concept PR:000045358 is used throughout the corpus, which denotes a family of proteins.
Even though the ID tagger produces correct labels, the \emph{spans-first} strategy used for PR gives precedence to the dictionary predictions in these cases.

\section*{Conclusions}

In this work, we present a concept-recognition architecture for parallel NER and NEN.
Compared to a sequential NER+NEN pipeline, our approach avoids error propagation from the span-detection to the normalisation step.
Modeling NEN as a sequence-labeling task allows it to operate directly on running text, at the cost of restricting the label set of the normaliser to the concepts from the training set.
We counter these limitations by fusing its predictions with the output of a span detector and a knowledge-based concept recogniser.

In the CRAFT shared task and in the current study, we have shown that parallel concept recognition can outperform a pipeline system created specifically for the CRAFT corpus.
Merging the predictions of a span and an ID tagger is a fruitful way of combining the complementary strengths of both of them.
However, the specifics of interpolating between span and ID predictions is subject to further research.
We took an empiric approach to pick the best harmonisation strategy for each annotation set.

For future work, we intend to test our approach on other datasets.
Even though the CRAFT corpus allows validating systems on a broad range of entity types, there is only little opportunity for direct comparison to competing approaches, as only few have published results for the latest versions (3 and 4) of CRAFT.

\subsection*{Availability of data and materials}
The code and configuration files created for conducting the experiments in the current study are hosted on GitHub, \url{https://github.com/OntoGene/craft-st}.
The trained models for the final results are available from Zenodo, \url{https://zenodo.org/record/3822363}.

\subsection*{Funding}
This work was supported by
the Swiss National Science Foundation [CR30I1 162758];  % Melanobase
and the Swiss Innovation Agency (InnoSuisse) [25587.2 PFES-ES].  % MedMon

\subsection*{Authors' contributions}
LF conducted the experiments and was a major contributor in writing the manuscript.
JC implemented the BioBERT-based system.
FR supervised the work and provided guidance on experiments.
All authors read and approved the final manuscript.

\subsection*{Acknowledgements}
We would like to thank the organisers of the CRAFT shared task 2019 for the well-organised competition with high-quality annotations and prompt support.

\bibliographystyle{unsrt}
\bibliography{refs}

\section*{Ancillary Files}
\subsection*{harmonisation-effect.csv}
Tabular version of the values presented in Figure~\ref{fig:ablation}.

\subsection*{hyperparameter-tuning.csv}
Full results of the hyperparameter-tuning process, performed over the training set in 6-fold cross-validation.

\end{document}